\documentclass[conference]{IEEEtran}

\usepackage{graphicx} 
\usepackage{subcaption} 
\usepackage{amsmath} 
\usepackage{amssymb}  
\usepackage[noend]{algpseudocode}
\usepackage{algorithm, multicol}
\usepackage{cancel}
\usepackage{siunitx}
\usepackage{xcolor}
\usepackage[numbers]{natbib}
\usepackage{hyperref}
\usepackage{comment}
\usepackage[normalem]{ulem}
\usepackage{booktabs}
\usepackage{multirow}
\usepackage{float}

\usepackage{times}

\usepackage{amssymb}
\usepackage{amsmath}
\usepackage{graphicx}
\usepackage{dblfloatfix}
\usepackage{xcolor}
\usepackage{verbatim}

\usepackage[font=small,skip=1pt]{caption}

\begin{document}

\title{\LARGE \bf
Learning Deep Parameterized Skills from Demonstration for Re-targetable Visuomotor Control}


\thanks{* denotes equal contribution}



%
\author{\authorblockN{Nishanth Kumar\authorrefmark{1}\authorrefmark{2},
Jonathan Chang\authorrefmark{1}\authorrefmark{2},
Sean Hastings\authorrefmark{2}, 
Aaron Gokaslan\authorrefmark{2}, \\
Diego Romeres\authorrefmark{3},
Devesh Jha\authorrefmark{3},
Daniel Nikovski\authorrefmark{3}, 
George Konidaris\authorrefmark{2} and 
Stefanie Tellex\authorrefmark{2}}
\authorblockA{\authorrefmark{2}Department of Computer Science\\
Brown University,
Providence, RI, 02912\\ Email: nkumar12@cs.brown.edu}
\authorblockA{\authorrefmark{3}Mitsubishi Electric Research Labs\\}
\authorblockA{\authorrefmark{1} denotes equal contribution}}

\maketitle

\begin{abstract}
  Robots need to learn skills that can not only generalize across similar problems, but also be directed to a specific goal. Previous methods either train a new skill for every different goal or do not infer the specific target in the presence of multiple goals from visual data. We introduce an end-to-end method that represents targetable visuomotor skills as a parameterized neural network policy. By training on an informative subset of available goals with the associated target parameters, we are able to learn a policy that can zero-shot generalize to previously unseen goals. We evaluate our method in a representative 2D simulation of a button grid and on both button-pressing and peg-insertion tasks on two different physical arms. We demonstrate that our model trained on 33\% of the possible goals is able to generalize to more than 90\% of the targets in the scene, for both simulation and robot experiments. Our method enables physical robots to learn parameterized policies to reliably insert a round peg into a targeted hole.

\end{abstract}

\IEEEpeerreviewmaketitle

\section{Introduction}
\label{sec:intro}
Recent advances have allowed for deep learning of visuomotor skills from demonstrations \citep{CodevillaSelfDriving, Zhang2017DeepIL}. However, because such skills are based only on visual input, they are insufficiently flexible for tasks with multiple potential goals visible in the same scene. For example, an intelligent robot should be able to identify and flip the correct light switch even in the presence of other light switches in the scene; it should be able to pick up the white queen on a chessboard even when there are other pieces, or press a specific button on a remote control. To achieve this behavior, the agent must learn a targetable skill: one that can be instructed to pick up \emph{this} piece, or flip \emph{that} switch, by taking as additional input a parameter vector that disambiguates its goal or some relevant aspect of the particular task at hand. An important question is then how to build targetable visuomotor skills in a sample-efficient manner.
   \begin{figure}
      \centering
      \includegraphics[width=\columnwidth]{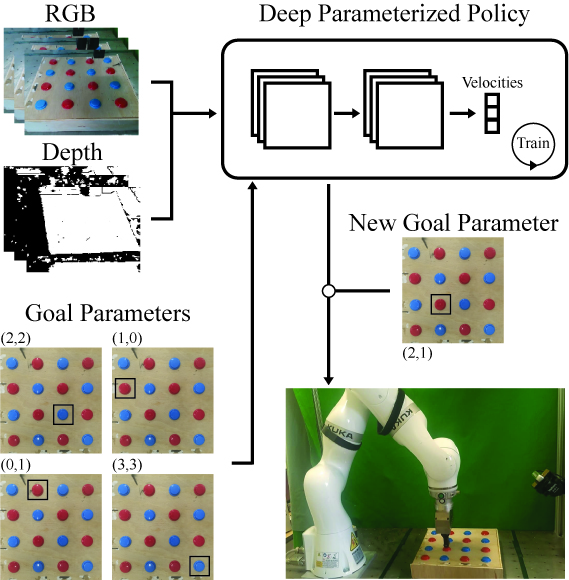}
      \caption{An overview of our pipeline. A deep neural network is trained with example goal parameters and raw sensor data, and evaluated on settings with unseen goal parameters. This figure shows the use of row/column indices on the button panel as the goal parameterization $\tau$.}
      \label{fig:overview}
   \end{figure}
   
While effective at learning an end-to-end visuomotor policy with a single goal from pixel data, current methodologies cannot be instructed either to a specific goal or towards a target not seen in training. Some approaches to solve this have involved either training a unique policy for different target behaviors \citep{CodevillaSelfDriving}, or creating a learning algorithm that could quickly converge to the desired policy given a few training examples \citep{duanoneshot, finnmetalearning, oneshotmetalearning}. However, the amount of training data required becomes prohibitive for related tasks such as pressing an elevator button, where one would need to train a separate skill for each button. Additionally, even though few-shot learning and meta-learning algorithms have drastically reduced the amount of training data needed, these algorithms require the system to have a separate policy to target a specific goal. Alternatively, \citet{dasilvaskills} introduced parameterized skills that allowed robot agents to target specific goals with a single policy; however, these methods were not end-to-end and used a separate module to infer the goal from pixel data. Furthermore, they exploited hand-selected features and did not learn directly from robot sensor data.

To address these shortcomings, we present a method to train targetable visuomotor skills from expert demonstrations. Our model consists of three modules: vision, auxiliary task, and control. The vision module takes as input an RGB and depth image and outputs a visual encoding. The auxiliary task module takes the visual encoding and the goal-parameter and tries to predict the final pose of the end-effector. Finally, the control module infers the next time step's linear velocity from the visual encoding, goal-parameter, and the predicted final state of the end-effector. By training on an informative subset of goal-parameters, we can train one policy that can retarget its learned visuomotor skill to previously-unseen goals.

We evaluate our method via a series of experiments in simulation and on two different robotic arms. In simulation, we use a two dimensional grid representing a $3\times3$ button panel to study our method's ability to generalize to novel goals. We show that the policy conditioned on the goal allows for the agent to generalize better in this multi-goal setting even when the button panel is moved arbitrarily during training. Moreover, we also demonstrate how conditioning with an informative representation of the goal further improves performance. Finally, we extend our algorithm such that it can generalize to novel variations of the goals in the scene. We then evaluate our method on two real-world robotic tasks that require significant precision in a continuous space. The KUKA LBR iiwa-7 and the MELFA RV-4FL arms were trained to solve button-pushing and peg-insertion tasks, respectively, on grid-based goals. In all of our experiments, our method could target specific goals and generalize to more than 90\% of goals after being trained on a third of the goals within the grid. We also demonstrate a goal-representation that allows our model to generalize to non grid-based goals with no additional training data and discuss how the choice of representation impacts learned skills.

\begin{figure}[t]
    \centering
    \begin{subfigure}[t]{0.5\columnwidth}
        \centering
        \includegraphics[height=1.4in]{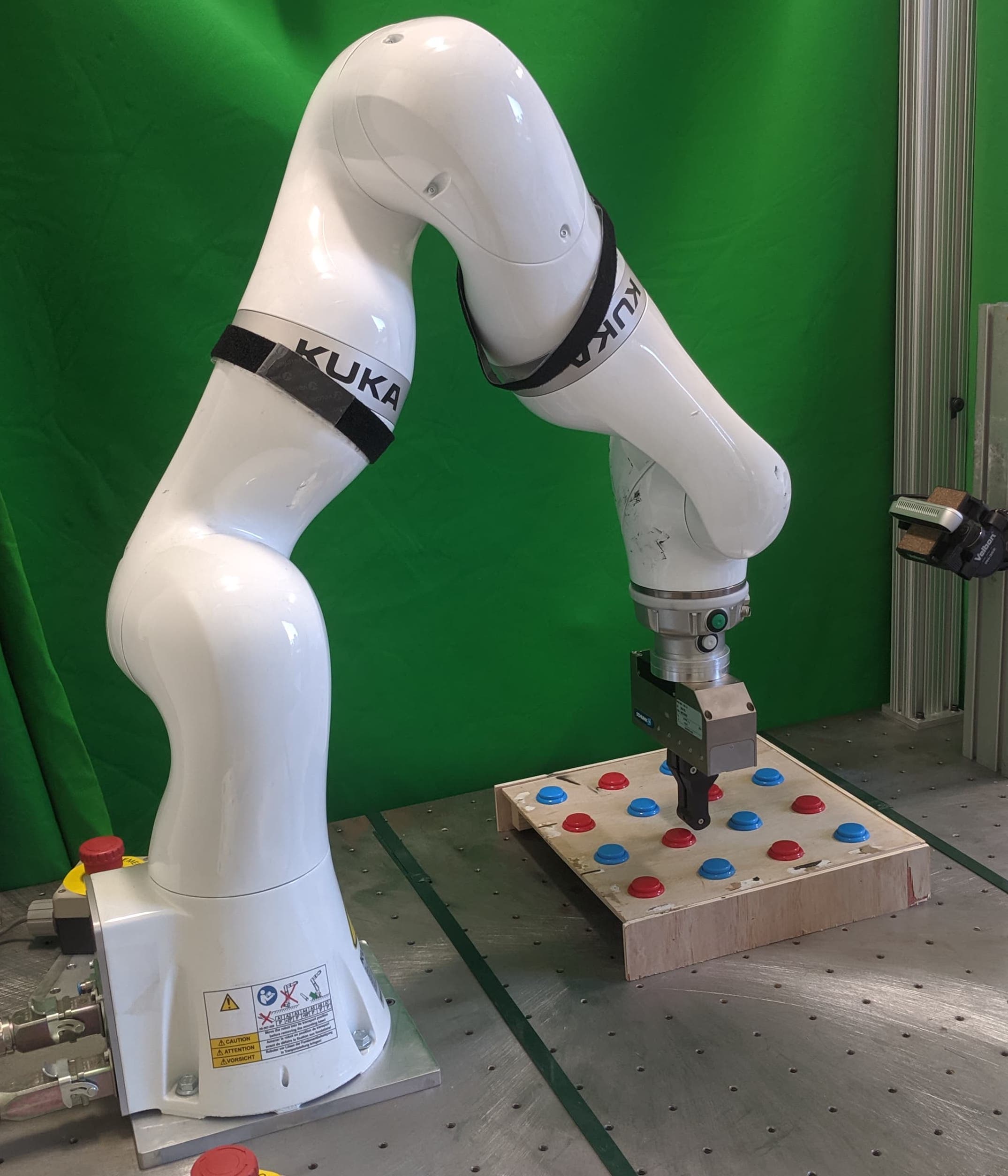}
        \caption{Button pressing}
        \label{fig:Kuka}
    \end{subfigure}%
    ~ 
    \begin{subfigure}[t]{0.5\columnwidth}
        \centering
        \includegraphics[height=1.4in]{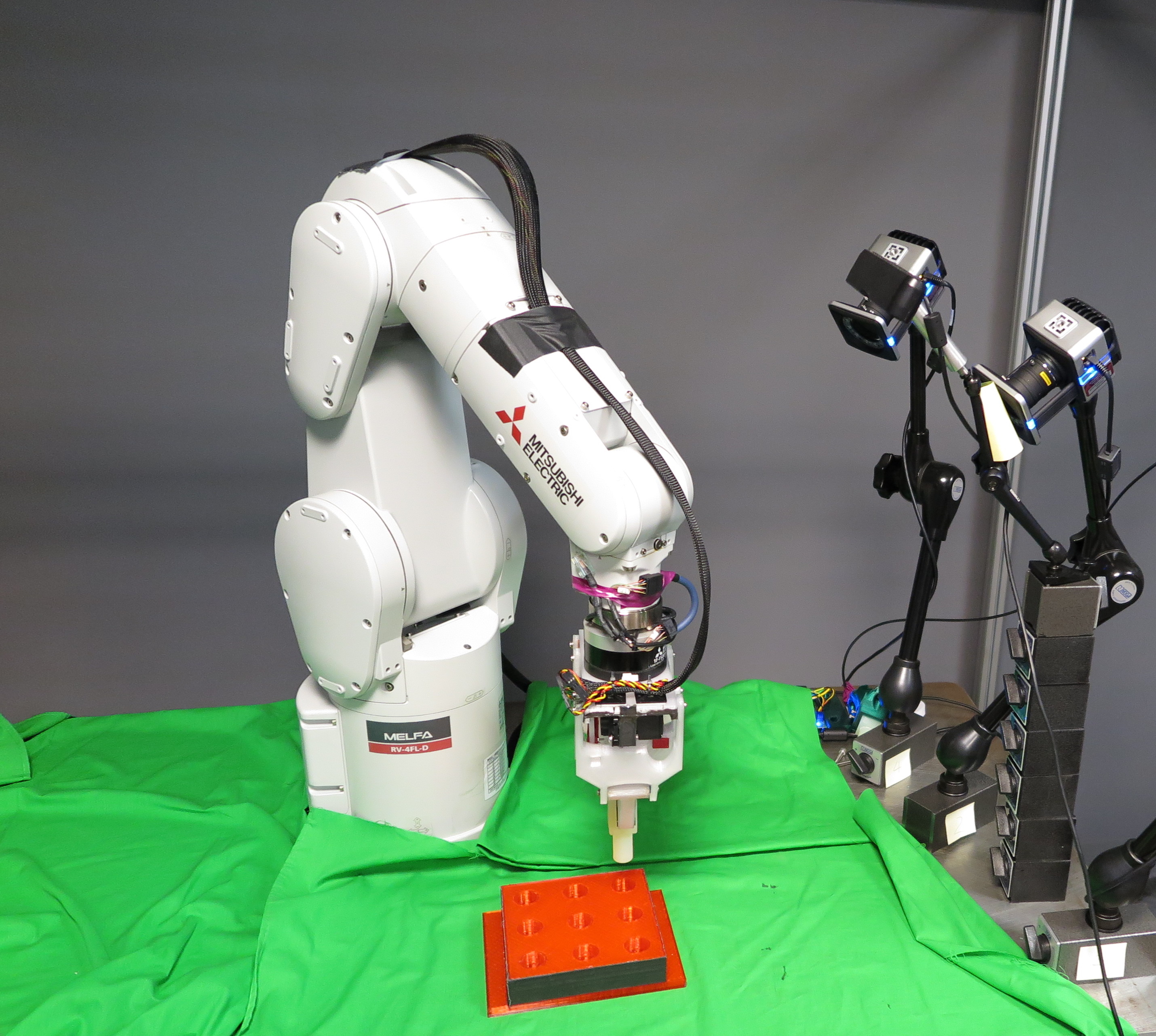}
        \caption{Peg insertion}
        \label{fig:MERLBot}
    \end{subfigure}
    \caption{(a) shows the KUKA LBR iiwa-7 with the button panel. (b) shows the MELFA RV-4FL robot with the peg-and-hole grid.}
\end{figure}

\begin{figure*}[thpb]
    \centering
    \includegraphics[width=\textwidth]{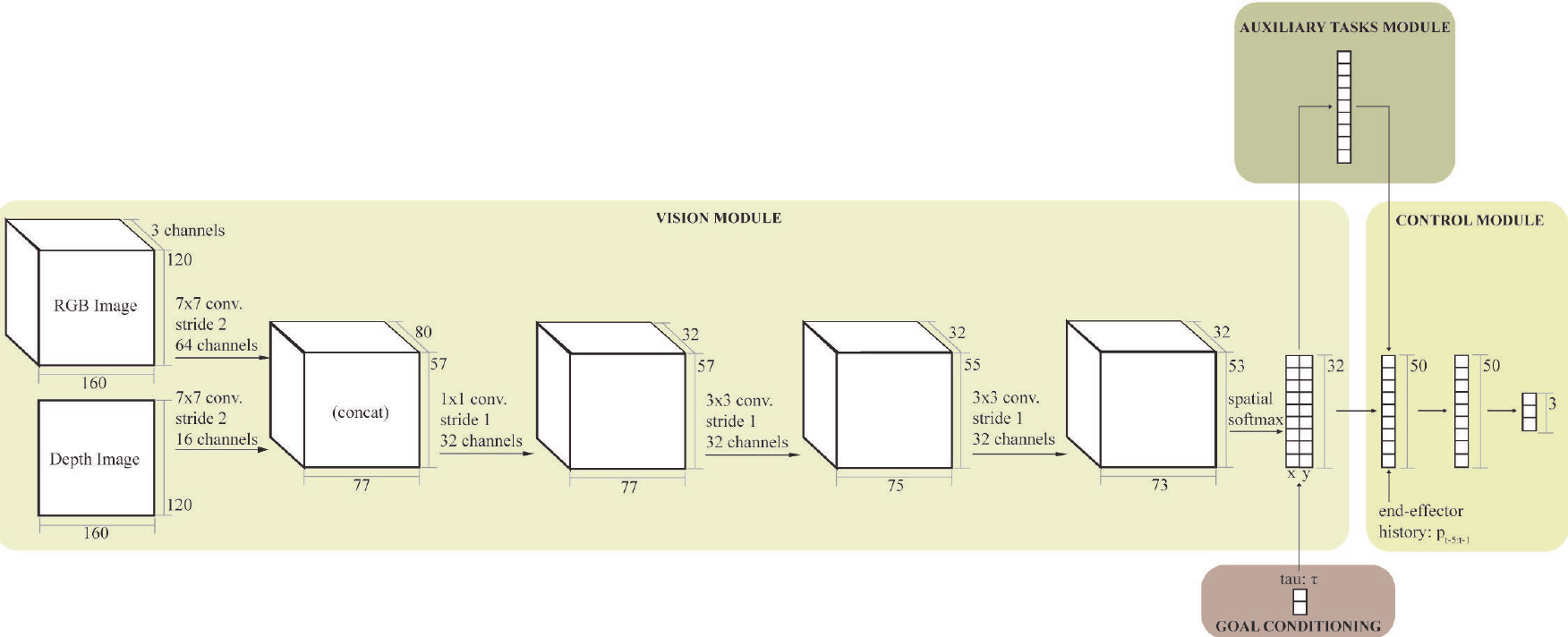}
    \caption{Architecture for the goal-parameterized deep imitation learning network. The vertical arrows indicate concatenation of the layer outputs and the other vectors.}
    \label{fig:architecture}
\end{figure*}
\section{Related Work}
\label{sec:related}
Our approach builds upon a few existing branches of work in learning from demonstration. In particular, our method combines the notion of parameterized skills \citep{dasilvaskills, brunorobot} with end-to-end deep visuomotor skills from demonstration \citep{endtoendvisuomotorpolicies, Zhang2017DeepIL}. We also take inspiration from several recent works that discuss \textit{goal-conditioning} for imitation and reinforcement learning.

Parameterized skill learning as described by \citet{dasilvaskills} aims to learn a mapping from a given task parameter vector to a policy \citep{dasilvaskills, brunorobot, deisenrothmultitask, duanoneshot, koberparameterizedrl, kupcsik, pastorparameterizedlfd, stulp}. All of these methods show zero-shot generalization to unseen goal parameters in settings with hand-designed compact state descriptors. Unlike prior work, we focus on directly learning from raw robot sensory data model-free instead of using hand-selected features or a simplified model of the environment. Like these works, our method can generalize to similar tasks with different goals by sharing data across tasks. Other approaches have achieved similar generalization through coupled dictionary learning \citep{dictionarycoupling}, gating functions \citep{gating}, and modular sub-networks \citep{devinmodularzeroshot}. However, unlike our approach, these methods are either unable to learn an end-to-end policy from raw sensory data or did not demonstrate their parameterization capabilities outside of simulation.

Work in end-to-end deep visuomotor skill learning has attempted to use raw sensory data from human demonstrations to learn robot motor control policies for a wide range of tasks
\citep{spatialautoencoder, endtoendvisuomotorpolicies, levinehandeye, rlwithnorewardengineering, vecerik, Zhang2017DeepIL, visualattention, CodevillaSelfDriving}. Our model specifically takes inspiration from the behavioral cloning algorithm proposed by \citet{Zhang2017DeepIL} which learns to map input RGB/depth sensor data to subsequent linear and angular velocities for a robot's end-effector. While these methods could generalize to unseen variations of the task with one possible goal in the visual frame, they fail to generalize in the presence of multiple  targets. For example, the behavioral cloning algorithm presented in \citet{Zhang2017DeepIL} would require 30 minutes worth of demonstrations for each button resulting in 270 minutes worth of demonstrations for nine buttons. Furthermore, since the skills learned are not parameterized by a task or goal parameter, a \textit{different skill} must be trained and selected for each variation of the task. However, our approach is able to train \textit{one} skill that can be varied depending on the task.

A similar approach to ours that also integrates parameterized skill learning and deep visuomotor skill learning was recently proposed by \citet{CodevillaSelfDriving}. Their method of \textit{conditional imitation learning} trains a deep visuomotor skill that varies with a \textit{command input} provided to the network in addition to the sensory inputs. However, their command input is used effectively only as a selector between different previously-trained behaviors. Specifically, they provided strings such as 'left', 'right' and 'straight' as command inputs to change the behavior of an autonomous vehicle. By contrast, our method can use the input goal-parameter to not only select previously-trained behaviors, but also smoothly generalize to novel instances of the behavior. In this way, our work can be seen as an extension of conditional imitation learning. Furthermore, our network architecture and application domain are significantly different to those of \citet{CodevillaSelfDriving}.

Ideas within goal-based Reinforcement Learning are also similar to those behind our work. Prior work has focused on learning a policy $\pi(a|s,g)$ that outputs actions conditioned both on the current state and goal state that maximizes the probability of reaching the goal quickly \citep{Schaul2015UniversalVF, GhoshActReprGoalConditioning, HeldAutomaticGoalGeneration, DBLP:GoalGAIL}. However, these methods are only applicable when the goal is a state within the agent's state-space. By contrast, our method is applicable to situations where the goal is not necessarily a state and may be described arbitrarily. Additionally, even though our evaluation and discussion focus on \textit{goal-parameterization} for robot learning, our method can handle more general parameterizations of a task. For instance, our method can support parameterizations that correspond to physical aspects of a task and enable transfer across tasks as in \citet{YuUniversalPolicyOIS}.

\section{Learning Deep Parameterized Skills}
\label{sec:learning}
This section presents a behavioral cloning algorithm that learns parameterized neural network policies. Given a set of skills $\mathcal{S}$, let $T = \{\tau^{(s)}\}$ be the set of goal parameters for each specific instance of the skill $s \in \mathcal{S}$ and $\Omega = \{(o^{(s)}_t, \tau^{(s)}, u^{(s)}_t)\}$ be the dataset that consists of sets of state observations, corresponding goal parameters, and controls collected for all $s\in\mathcal{S}$. Now let $\pi_{\theta}(u_t|o_{t}, \tau)$ be our neural network policy, parameterized by $\theta$, that learns control policies that are dependent on our goal parameters, $\tau$. Our neural network architecture closely follows that of \citet{Zhang2017DeepIL} with the following modifications.

\subsection{Task Parameterization}
As proposed by \citet{dasilvaskills}, a parameterized skill is a multi-task policy that maps input task parameters to end-effector controls. We learn the mapping $T \rightarrow \theta_{\mathcal{S}}$ where $T$ is the set of task parameters and $\theta_{\mathcal{S}}$ is the multi-task policy parameters for the family of skills $\mathcal{S}$. We define $\tau^{(s)} \in T$ as the vector of task parameters injected into our policy for skill $s\in\mathcal{S}$. In our particular tasks, $\tau$ describes information directly related to the goal, and we henceforth refer to it as the goal parameter, though it does not need to be a goal in general.

\subsection{Neural Network Policies}
The neural network takes raw sensory data and a goal parameter as input and outputs robot motions. More formally, for time step $t$, the inputs $o_{t} = (I_{t}, D_{t}, P_{t-5:t-1})$ include (1) RGB images $I_t \in \mathbb{R}^{160\times120\times3}$, (2) depth images $D_t\in\mathbb{R}^{160\times120}$, and (3) positions of the end-effector $P_t\in\mathbb{R}^{3}$ for the 5 most recent steps where $P_{t-5:t-1} \in\mathbb{R}^{15}$. Furthermore, the network also takes as input $\tau$ which is the goal parameter.

\begin{figure}[thpb]
    \centering
    \includegraphics[width=\columnwidth]{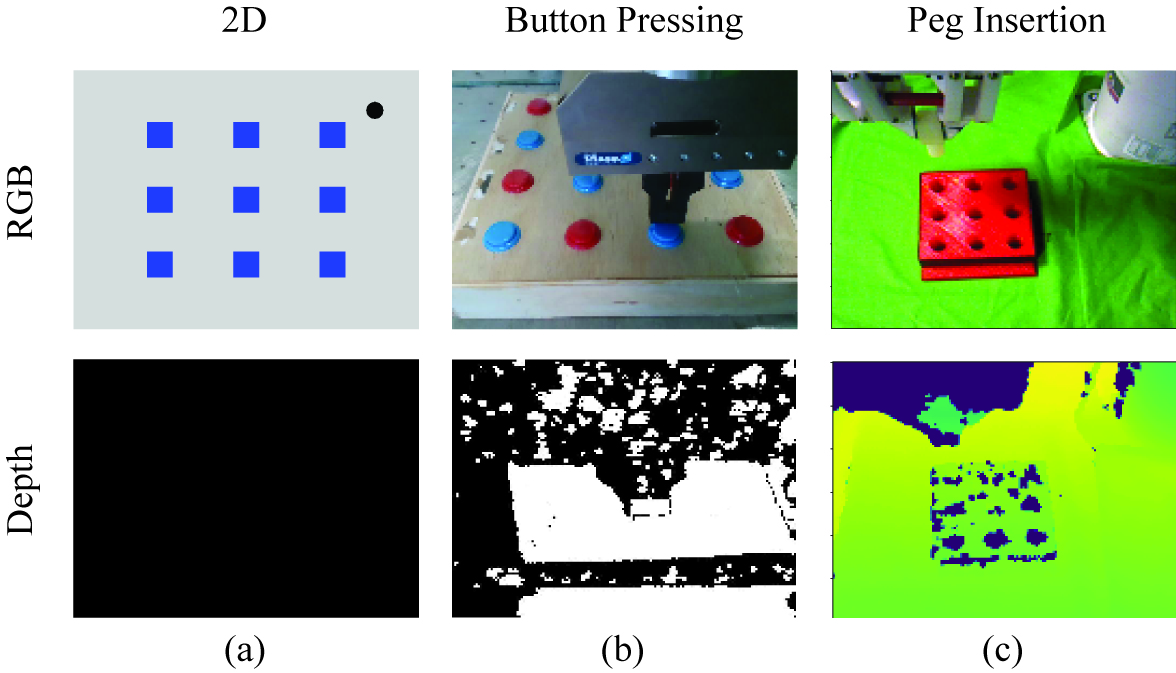}
    \caption{The first and second row show the RGB and depth image inputs respectively for the (a) 2D, (b) button pressing, and (c) peg insertion experiments.}
    \label{fig:inputs}
\end{figure}

Given these inputs, the neural network outputs the current control $u_t$ described as the linear velocity, $v_t\in\mathbb{R}^3$ of the end-effector.

The neural network architecture can be decomposed into three modules, $\theta = (\theta_{vision}, \theta_{aux}, \theta_{control})$. The first module consists of convolutional layers with a spatial-softmax layer \citep{spatialautoencoder, endtoendvisuomotorpolicies} to extract spatial feature points from the image to generate the current state encoding (Eq. \ref{eqn:cnn}). Every convolutional layer is followed by a layer of rectified linear units \citep{nairRELU} for nonlinearity:
\begin{equation}
    \label{eqn:cnn}
    f_t = \textrm{CNN}(I_t, D_t; \theta_{vision}).
\end{equation}
This is followed by a small fully connected network that takes as input the state encoding, $f_t$, and the goal parameter, $\tau$, to predict the auxiliary task:
\begin{equation}
    \label{eqn:aux}
    a_t = \textrm{NN}(f_t, \tau; \theta_{aux}).
\end{equation}

Finally, given the extracted state encoding $f_{t}$, end-effector positions $P_{t-5,t-1}$, goal parameters $\tau$, and auxiliary prediction $a_t$, we use a fully-connected network to predict the current time step's controls:
\begin{equation}
    \label{eqn:control}
    u_t = \textrm{NN}(f_t, P_{t-5:t-1}, \tau, a_t; \theta_{control}).
\end{equation}

After each non-output fully-connected layer, we add a layer of rectified linear units \citep{nairRELU} for nonlinearity.

\subsection{Auxiliary Prediction Tasks}
Our network includes an auxiliary prediction task as another means of self-supervision, resembling the approach taken by \citet{Zhang2017DeepIL}, who found that leveraging the extra self-supervisory signals resulted in increased data efficiency. As mentioned in \citet{Zhang2017DeepIL}, there could be multiple auxiliary tasks, but we found that for our tested tasks, one auxiliary task module was sufficient. We added a module of two fully-connected layers after the spatial-softmax layer (Eq. \ref{eqn:aux}) and fed the final layer of these modules back into the control module (Eq. \ref{eqn:control}) as shown in Figure \ref{fig:architecture}.

The auxiliary tasks that we chose were limited to using the information that could be inferred from the dataset $\Omega$. For our experiments, we had an auxiliary module for predicting the final end-effector pose for the task. We also found that generalization across novel goal parameters improved with the goal parameters being fed into this auxiliary prediction. All of these auxiliary prediction tasks were trained concurrently with the rest of the network.

\subsection{Loss Functions}
The loss function used for learning is the commonly used behavioral cloning loss also used by \citet{Zhang2017DeepIL}. This approach involves using a linear combination of four loss terms to guide different aspects of the output. First both $l1$ and $l2$ losses are used to directly fit to the training trajectories, as is done for most behavior cloning algorithms
. Given an example set of $(o_t, \tau, u_t)$, we have the losses:
\begin{align}
    \label{eqn:loss_l1}
    \mathcal{L}_{l1} &= ||\pi_{\theta}(o_{t}, \tau) - u_t||_1\\
    \label{eqn:loss_l2}
    \mathcal{L}_{l2} &= ||\pi_{\theta}(o_{t}, \tau) - u_t||^2_2.
\end{align}
Furthermore, there is an arc-cosine loss to enforce consistency between the directions of the output and target velocity, since we care more about matching the direction of the output velocity than its exact magnitude.
\begin{equation}
    \label{eqn:loss_cos}
    \mathcal{L}_{acos} = \arccos{\frac{u_t^T\pi_{\theta}(o_t, \tau)}{||u_t||_2||\pi_{\theta}(o_t, \tau)||_2}}.
\end{equation}
Finally, an $l2$ loss for the auxiliary prediction task:
\begin{equation}
    \label{eqn:loss_aux}
    \mathcal{L}_{aux} = ||\textrm{NN}(f_t;\theta_{aux}) - a_t||^2_2.
\end{equation}

The loss function for the whole algorithm is a weighted sum of the losses described by the above equations:
\begin{equation}
    \label{eqn:loss}
    \mathcal{L} = \lambda_{l1}\mathcal{L}_{l1} + \lambda_{l2}\mathcal{L}_{l2} +
    \lambda_{acos}\mathcal{L}_{acos} +
    \lambda_{aux}\mathcal{L}_{aux}.
\end{equation}

$\lambda = (\lambda_{l1}, \lambda_{l2}, \lambda_{acos}, \lambda_{aux})$ was tuned roughly for each domain so that the loss terms would have the same order of magnitude. For the button pressing task we use $\lambda = (1.0, 0.01, 0.005, 1.0)$, whereas for the peg-insertion task we use $\lambda = (1.0, 1.0, 10.0, 1.0)$. Policies were optimized using NovoGrad~\citep{novograd} with a learning rate of 0.0005 and batch size of 64. Training was done with randomly sampled batches from the dataset $\Omega$.
\section{Evaluation}
\label{sec:evaluation}

We conducted several experiments to study our method's ability to learn targetable, generalizable visuomotor skills in both simulated and real-world domains.

\subsection{ Core Experiments}

Our first set of experiments investigates whether a particular choice of representation for our goal-parameter $\tau$ improves the performance and generalization of behavior cloning. We chose not to perform ablation studies on the vision, control and auxiliary task modules because these are directly derived from \citet{Zhang2017DeepIL}, who perform extensive ablation studies on these factors.

\subsubsection{2D Button Simulation}
\label{subsubsec:2dsim}
We first experimented on a 2D simulation of the robot button-pressing task. We used a 3$\times$3 grid of blue squares representing buttons and a black circle representing the agent, shown in Figure \ref{fig:inputs}a. We designed the simulation such that the agent would occlude the squares when it passed over them.

\begin{figure}
    \centering
    \includegraphics[width=\columnwidth]{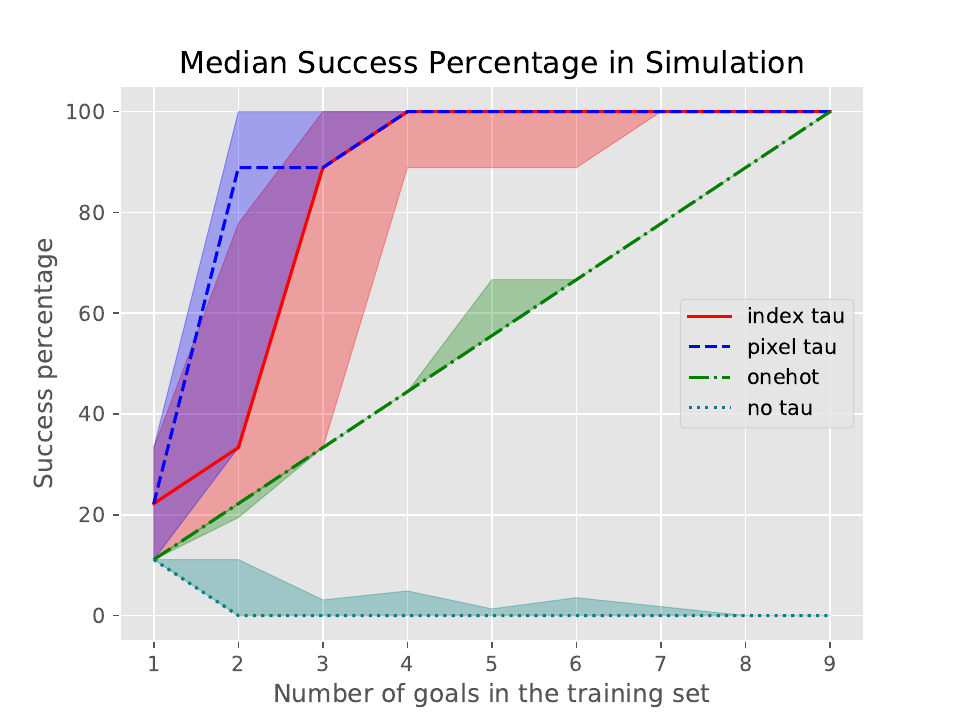}

\caption{Results from the 2D simulation domain of all $\tau$s used (including ablated $\tau$). Lines are median across random subsets, and shading shows the range across subsets.}

    \label{fig:2dResults}
\end{figure}

We collected 100 trajectories for each square where the agent began at a random position along the right and top edges of the scene. As shown in Figure \ref{fig:inputs}a, the depth image input was a black screen. For this experiment, we used the row/column index of the square as $\tau$. For example, we set $\tau$ as $(0,0)$ for the top-left square, $(2,0)$ for the bottom-left, and so on.

We trained our network on random subsets of $\tau$ corresponding to the nine squares to test how well our model generalized. As an example: $\{(0,0), (0,2), (2,0), (2,2)\}$ is a length $4$ subset of $\tau$'s corresponding to the $4$ corner squares in our grid. For every random subset, we trained our network for 50 epochs and evaluated the performance of our model on 100 trials for each button. The trajectories for the trials were generated by moving the agent in one of eight cardinal directions towards the goal at each time-step. A trial was counted as a success only if the agent slowed to a complete stop at the correct blue square.

We conducted an ablation study to evaluate the effect of our model's goal parameterization. Our ablation of the goal-parameter was equivalent to the architecture shown in Figure \ref{fig:architecture} without the goal-conditioning module, which is mostly equivalent to the architecture used by \citet{Zhang2017DeepIL}. In addition, we evaluated a version of our algorithm that uses a one-hot vector where a goal corresponded to a randomly-chosen index of a nine-dimensional vector. This representation is \textit{unstructured} because there is no information for the agent to exploit to generalize to previously-unseen goals. Thus, it should theoretically only be able to select amongst goals it has already seen during training and is similar to the \textit{conditional imitation learning} approach from \citet{CodevillaSelfDriving}, albeit with a different architecture that is more suited to this task. 

For our index-based and one-hot representations as well as our ablation, we repeated the above experiment involving training and evaluating on 100 trials of data with the specified model changes. The results are displayed in Figure \ref{fig:2dResults}.

Our experiments show that injecting the goal parameter $\tau$ defined by the relative row/column indices allows the original behavioral cloning algorithm to select specific goals and even generalize to unseen ones via interpolation. As Figure \ref{fig:2dResults} shows, the version of the network with the goal-conditioning ablated is only able to achieve approximately $11\%$ success rate when trained on only one square and its performance generally degraded with more squares in the training set probably because the network was simply averaging the trajectories it was trained on. We also see that the one-hot vector representation of the goal parameter only allows for the algorithm to successfully reach squares seen in training. The addition of our structured goal-parameter $\tau$ as the relative row-column index allows the network to learn the mapping from $\tau$ to the location of a square, and thus enable it to not only select squares it has seen during training, but also generalize to new squares based on novel $\tau$ inputs.

Additionally, we found that given this representation for $\tau$, the network was able to consistently achieve perfect performance after having trained on roughly $7/9$ of the possible targets regardless of the specific combinations of goal-parameters in the training set. We also found that given optimal selection of goal parameters in the training set, we were able to achieve full generalization for the whole button panel with $33\%$ of the goal-parameters represented in the training data, which is much more data-efficient than the one-hot baseline. Our experiments show that our network is able to generalize to the entire space of goals after seeing roughly a third of the possible goals, provided that the training set represents an informative subset of the entire goal-parameter space.

\subsubsection{Robot Button-Pressing Task}
\label{subsubsec:button-pressing}
In this experiment, we show that our method can work robustly on a real-world robotic task. We used a KUKA LBR iiwa-7 equipped with a Schunk gripper to press buttons on a 3D, $4\times4$ button panel. Similar to Section \ref{subsubsec:2dsim}, we parameterized our button-grid with a row/column tuple of the button's location on the grid. For training, we collected 100 trials of the robot's end-effector beginning at a random position and following a straight line with noise to the specified button. The end-effector's final position was varied with noise drawn from a Gaussian distribution such that the robot would press the button differently each time.

\begin{figure}[t]
    \centering
    \includegraphics[width=\columnwidth]{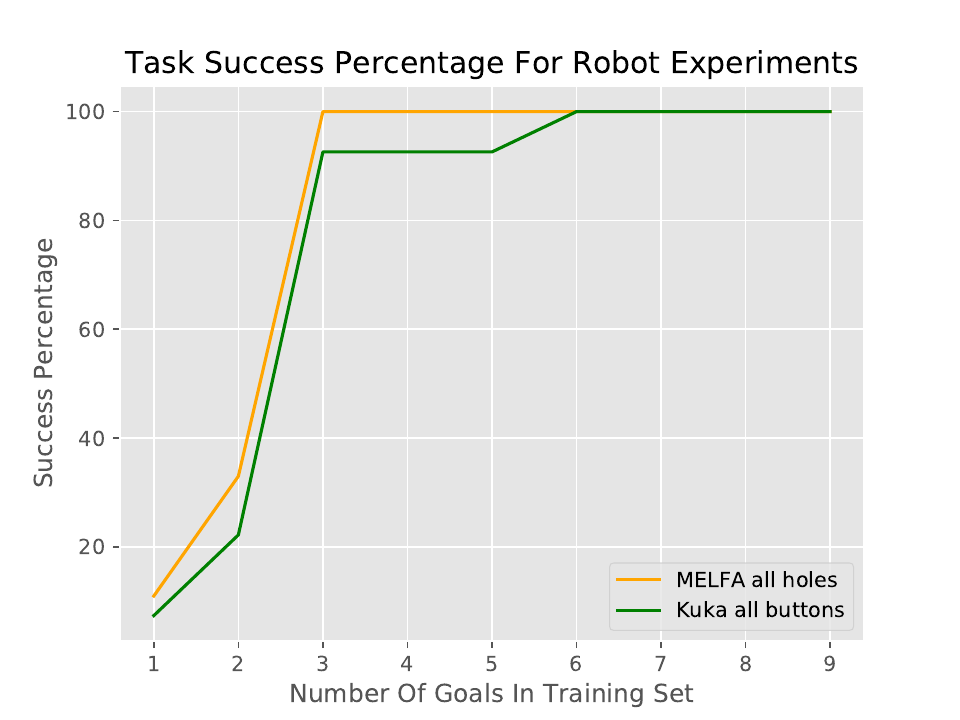}
    \caption{Results from our experiments on the MELFA and KUKA robots. }
    \label{fig:3dResults}
\end{figure}

For this experiment, we used specific subsets of buttons within a $3\times3$ section of the grid as training data. We chose combinations that had been found to generalize well in our two-dimensional simulation. We evaluated for three attempts on each button and deemed an attempt successful if the robot pressed the button. Results are displayed in Figure \ref{fig:3dResults}.

The performance of the robot on the task was similar to the average performance with the row/column $\tau$ for our 2D simulation. The robot always successfully pushed buttons seen during training. After having been trained on just three buttons, the robot successfully generalized to $88\%$ of previously-unseen buttons and $92\%$ of all buttons. Interestingly, the average performance of the robot stayed the same when trained on three to five buttons because the robot failed to press exactly two unseen buttons in each of these cases. However, we qualitatively observed that the robot did get progressively closer to succeeding when trained on more buttons, but was still not close enough to successfully press. When trained on six buttons or more, the robot achieved a $100\%$ success rate on pressing all buttons in the grid.

\subsubsection{Robot Peg-Insertion Task}

\begin{figure*}[t]
    \centering
    \begin{subfigure}[t]{\columnwidth}
        \centering
        \includegraphics[height=2.2in]{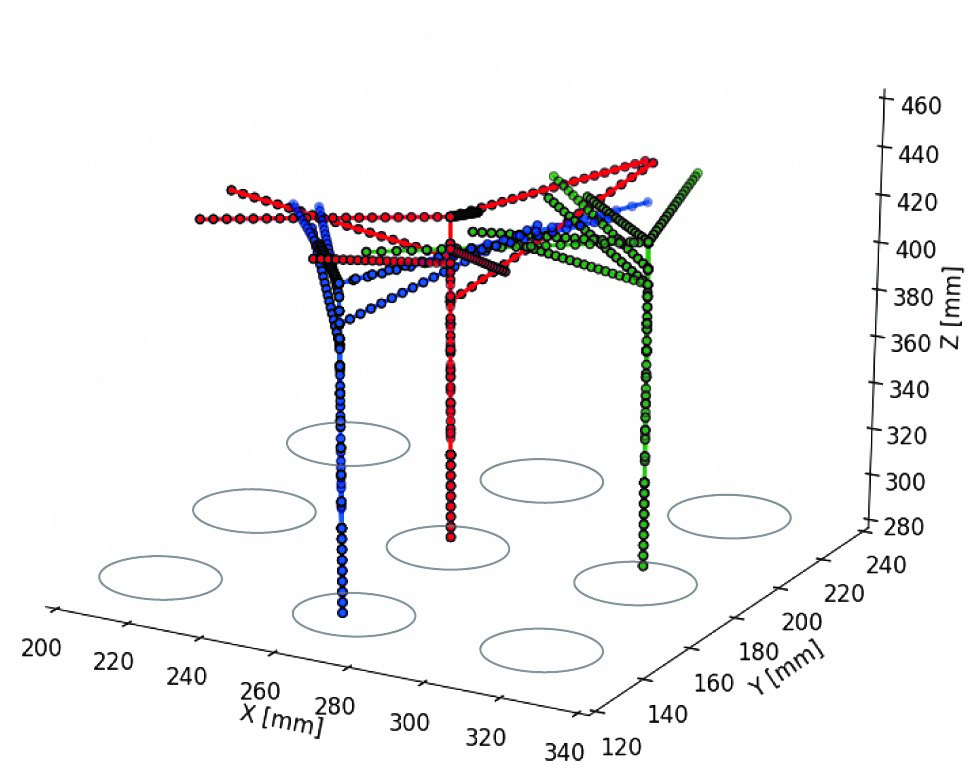}
        \caption{Trajectories in training set}
        \label{fig:traj_train}
    \end{subfigure}%
    ~ 
    \begin{subfigure}[t]{\columnwidth}
        \centering
        \includegraphics[height=2.2in]{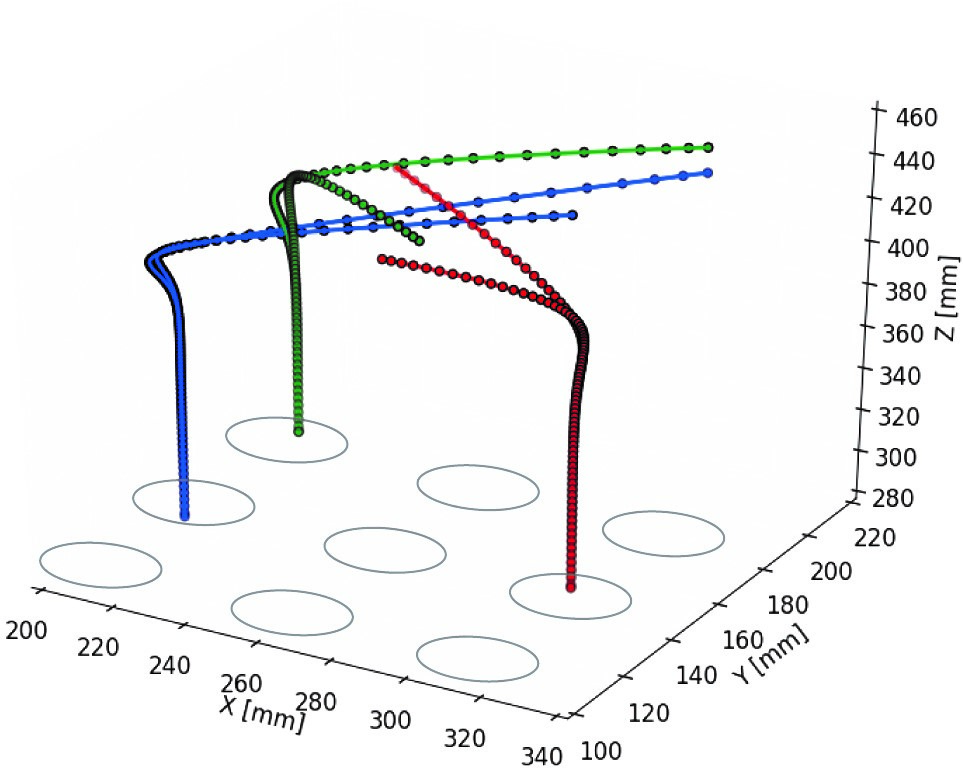}
        \caption{Trajectories from evaluation}
        \label{fig:traj_test}
    \end{subfigure}
    \caption{Plots of the trajectories that were used in training and produced in evaluation for the peg-insertion task by the MELFA RV-4FL. The 3x3 grid on the xy plane represent the various goals for this task.}
    \label{fig:trajexample}
\end{figure*}
We further evaluated our method on a real-world robotic task that required significant precision and was performed by a different robot arm. As pictured in Figure \ref{fig:MERLBot}, we used a MELFA RV-4FL robotic arm to perform peg-insertions in a 3$\times$3 grid of holes. We performed an experiment similar to the one in Section \ref{subsubsec:button-pressing}, with a different robot, task-setting, and subsets of goals chosen. The collected trajectories are straight lines from random start positions, uniformly sampled on the area above the holes grid, to a waypoint of random heights directly above the hole, and then a straight path downward into the hole to complete the insertion. It was critical that the robot learn a behavior where it inserts the peg vertically downward into the hole because the tolerance was less than $1.5$ cm. Shown in Figure \ref{fig:traj_train}, an example three-goal subset that we trained on was $\tau \in \{(1,1), (1,2), (2,1)\}$. During data collection 60 such trajectories were collected for each hole, for a total of 540 insertions. Results of successful insertions for different training combination of holes are displayed in Figure \ref{fig:3dResults}.

Our method performed remarkably well on this task. The robot generalized to all nine holes with a $100\%$ success rate after having seen insertions performed on only three holes during training. The end-to-end nature of our approach enabled us to learn a skill that not only predicted the positions of previously-unseen holes, but also varied the behavior to enable successful insertions despite a small tolerance. Additionally, we found that the outputted control trajectories to new goals were very smooth as shown in Figure \ref{fig:traj_test}. 
\begin{table}[b]
    \centering
    \begin{tabular}{cccc}
    \toprule
                & 2D & Button-Pressing & Peg-Insertion \\
                \midrule
        Indices & 3  & 6               & 3             \\
        Pixel   & 2  & 4               & -             \\
    \bottomrule
    \end{tabular}
    \caption{Lowest number of goal parameters, either row/column indices or pixel coordinates, at which we observed perfect 100\% generalization to the 3x3 grid of goals in our experiments.}
    \label{tab:summary}
\end{table}

The differences between our performance on the peg-insertion and button-pressing tasks can likely be explained by two differences: the noise in the training trajectories and the subsets of the goals that the robots were trained on. As shown in Figure \ref{fig:traj_train}, the training trajectories for our peg-insertion experiments had no noise on their final positions because the insertion tolerance was too small to induce much noise. However, the training trajectories for our button-pressing experiment had significant noise on the end-effector's final position. This could have lowered the precision of the model, leading to near-misses for buttons that were not seen during training. In addition, the two tasks did not use the same subsets of goals for training in every case. It is possible that some of the subsets used for the peg-insertion task were more optimal than those used for the button-pressing task.

\subsection{Investigating a different representation for $\tau$}


Our second set of experiments studies how the choice of representation for the goal-parameter can affect learned skills in the various different domains. Specifically, we chose to use the pixel-location of the goal as $\tau$ instead of the row/column indices. This is a particularly useful and intuitive representation within an end-to-end visuomotor learning system because a human user can choose a goal simply selecting a point in the robot's field of view. 

We first repeated our initial experiments on both the 2D simulation and button pressing tasks, but with the pixel location of the center of the goal button as the $\tau$ input. In general, we found that the results were similar to those obtained with the index representation. As shown in Figure \ref{fig:2dResults}, using this pixel representation for $\tau$ demonstrated performance comparable to, if not better than, the row/column based representation. We observed a similar result for the button pressing task: as can be seen in Table 1, this representation was able to achieve $100\%$ generalization to all buttons in the panel after training on only $4$ buttons as opposed to the $6$ buttons required by the index parameterization.

\begin{figure}[tp]
    \centering
    \begin{tabular}{ll}
        \begin{subfigure}[t]{0.45\columnwidth}
            \includegraphics[width=\columnwidth]{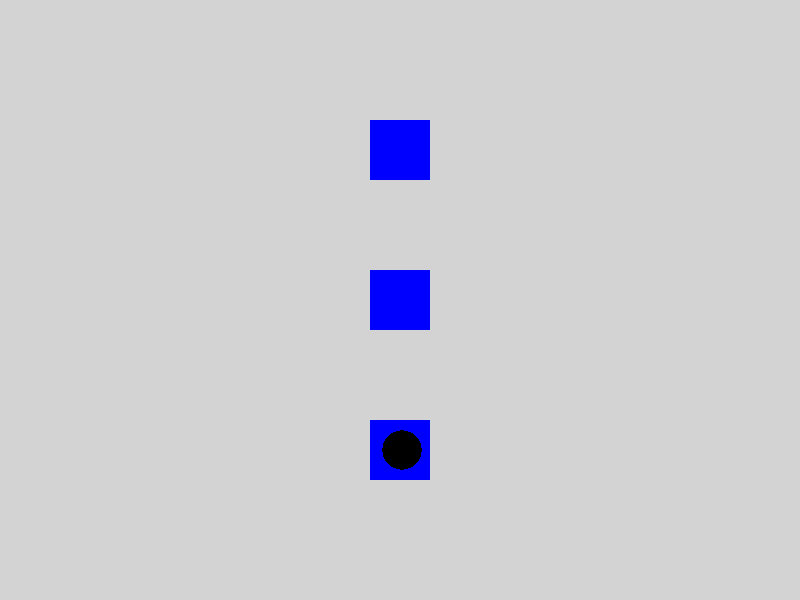}
            \caption{Column}
        \end{subfigure}
        &
        \begin{subfigure}[t]{0.45\columnwidth}
            \includegraphics[width=\columnwidth]{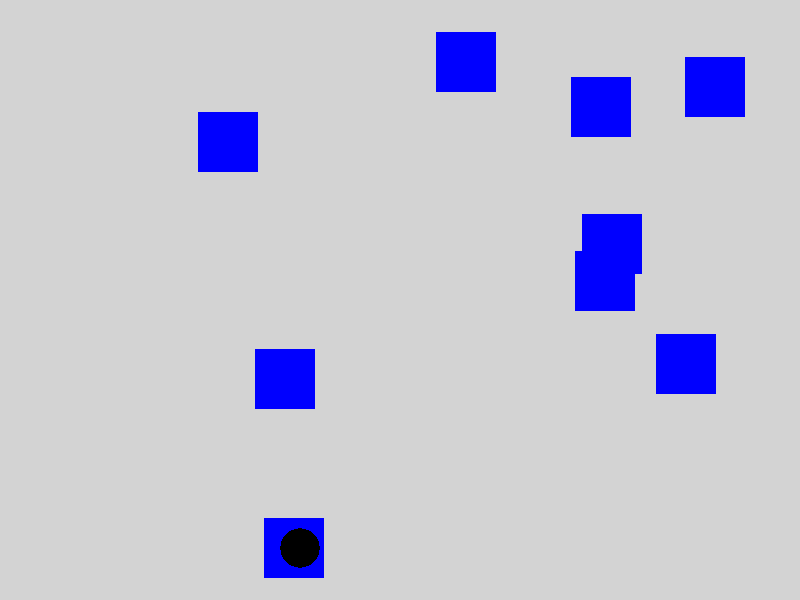}
            \caption{Scramble}
        \end{subfigure}
        \\
        \begin{subfigure}[t]{0.45\columnwidth}
            \includegraphics[width=\columnwidth]{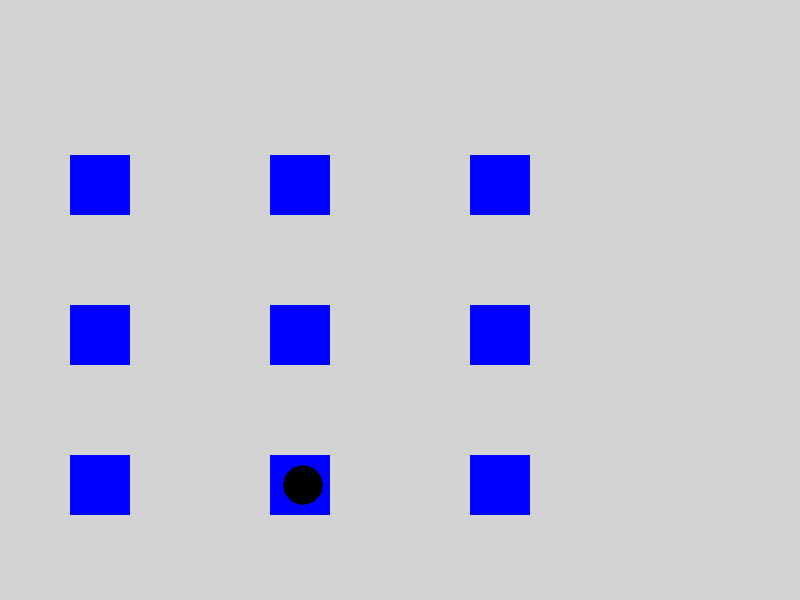}
            \caption{Shift}
        \end{subfigure}
        &
        \begin{subfigure}[t]{0.45\columnwidth}
            \includegraphics[width=\columnwidth]{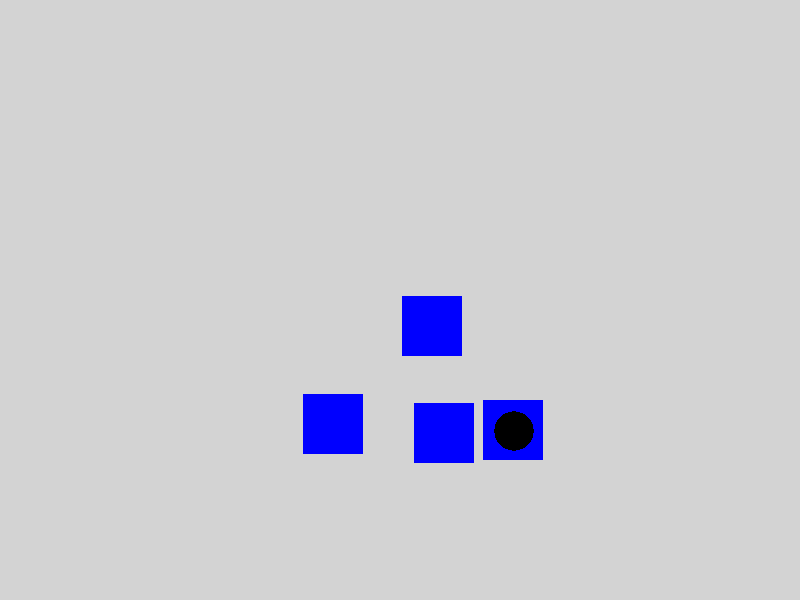}
            \caption{Clump}
        \end{subfigure}
    \end{tabular}
    \caption{Various button arrangements the model generalized to after being trained on original 3x3 grid using goal pixel location as $\tau$.}
    \label{fig:2dgen}
\end{figure}

\begin{figure}[b]
    \centering
    \begin{tabular}{ll}
        \begin{subfigure}[t]{0.45\columnwidth}
            \includegraphics[width=\columnwidth]{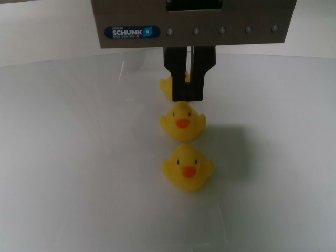}
            \caption{Column}
        \end{subfigure}
        &
        \begin{subfigure}[t]{0.45\columnwidth}
            \includegraphics[width=\columnwidth]{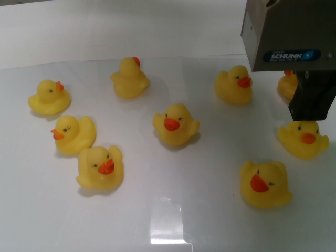}
            \caption{Scramble}
        \end{subfigure}
        \\
        \begin{subfigure}[t]{0.45\columnwidth}
            \includegraphics[width=\columnwidth]{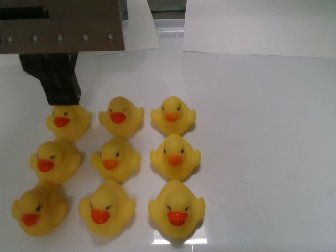}
            \caption{Shift}
        \end{subfigure}
        &
        \begin{subfigure}[t]{0.45\columnwidth}
            \includegraphics[width=\columnwidth]{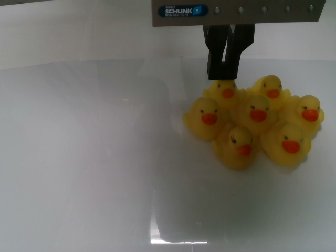}
            \caption{Clump}
        \end{subfigure}
    \end{tabular}
    \caption{Various goal arrangements the model generalized to after being trained on the original 3x3 button grid using goal pixel location as $\tau$.}
    \label{fig:3dgen}
\end{figure}

However, unlike with the index parameterization, using the pixel parameterization allowed skills to generalize to a wide variety of novel button arrangements as shown in Figure \ref{fig:2dgen} for the 2D simulation. For the robot button-pressing tasks, we discovered the button-pressing skill generalized to a situation with completely different visual input where rubber ducks were used instead of buttons as shown in Figure \ref{fig:3dgen}. The generalization in button-pressing task is particularly surprising because the task is so visually different from the button board that the robot was trained on.

This surprising result can be explained by realizing that the visual input is not necessary for the robot to accomplish the skill. The robot simply needs to learn a function mapping the pixel coordinate $\tau$ to the corresponding position in 2D or 3D space depending on the task and output velocities to move in that direction. Since the visual input is kept constant during training, and changing the input $\tau$ directly corresponds to a change in the agent's goal position \textit{regardless of the visual input}, the function to be learned is rather simple. In this way, using the pixel representation for $\tau$ for these tasks completely invalidates the need for visual input to learn the relevant skills.

In order to contrast this result against a harder task that requires significant use of vision to solve, we performed experiments on a modified version of our 2D simulation domain. We provided the network with the index based $\tau$ as input and in each training trajectory, the button panel grid was placed randomly within the screen (similar to the \textit{shift} arrangement depicted in Figure 8.c). This meant that the network had to use the vision module to localize the button panel in order to make use of the relative positional information in $\tau$.

In simulation, our model's performance on this task was comparable to the performance achieved on the simulated stationary panel task using the row/column index parameterization. The performance of the model was qualitatively unaffected by the mildly increased complexity of the task, and the vision module is therefore shown to be able to localize the goals under slightly varied visual conditions. Furthermore, this skill was not able to transfer to button arrangements shown in Figure 8 a, b and d, since the $\tau$ input and visual information alone are not enough to complete the skill.

We ran this simulation experiment first with the same data quantity as the original experiment, then ran it again but with only 50\% of the original data (50 trajectories per button). Trained on 4 buttons for 200 total trajectories, we were able to generalize to the entire button board, and any board placement. This is the same number of training trajectories used by \citet{Zhang2017DeepIL} to achieve reaching to a goal object in an arbitrary position on a table. By comparison, we use a similar amount of data to achieve a slightly more complex task (button pressing vs reaching), and we also generalize across sub-tasks (which button to press). Furthermore, the fact that we were able to succeed on generalizing within this task with so few trajectories suggests that the tasks attempted within section IV A. probably require a much smaller amount of trajectories than collected to completely learn those tasks.

These experiments illustrate how changing the representation for $\tau$ can drastically affect the learned skill. Using the pixel location as $\tau$ effectively allowed the network to learn a skill that did not depend on the visual input and thus could transfer to vastly different scene variations from the original task. On the other hand, using the row/column index as $\tau$ with a moving button panel required the learned skills to be dependent on the visual input and thus unable to transfer to different scene variations. Thus, different tasks could warrant completely different choices of representation for $\tau$ depending on how easy to learn and generalizable a user wants the  skill to be. Studying various choices of representation for $\tau$ on a wide variety of different, difficult tasks is a promising direction for future work.

\section{Conclusion}
\label{sec:conclusion}
We introduce a method that uses a neural network to learn deep parameterized visuomotor skills. We show that our method can learn to perform new instances of a task that were not seen at training time and demonstrate this via experiments on different tasks in a 2D simulation and on two different physical robots. We empirically study our method's generalization and dependence on user-specified goal parameters and show that it is able to generalize to all possible instances of various tasks after having seen at most six out of nine instances at training, provided a diverse subset of goals. This is summarized in Table \ref{tab:summary}. We also show that depending on the choice of goal parameter representation, our model can generalize to different scene variations of multiple goals with no additional training data.


\bibliographystyle{plainnat}
\bibliography{references}

\end{document}